\documentclass{elsarticle}

\makeatletter
\def\ps@pprintTitle{%
 \let\@oddhead\@empty
 \let\@evenhead\@empty
 \def\@oddfoot{}%
 \let\@evenfoot\@oddfoot}
\makeatother

\usepackage[hidelinks]{hyperref}
\usepackage{hyperref}
\usepackage[utf8]{inputenc}
\usepackage{rotating}
\usepackage{subfig}
\usepackage{siunitx}
\usepackage{amsmath}
\usepackage{amssymb}
\usepackage{amsthm}
\usepackage{numprint}
\usepackage{graphicx}
\usepackage{multirow}
\usepackage{caption}
\usepackage{subcaption}
\usepackage{booktabs}
\usepackage{tabularx}
\usepackage{array}
\usepackage[T1]{fontenc}
\usepackage{gensymb}
\usepackage{dirtytalk}
\usepackage{tablefootnote}
\usepackage{bm}
\usepackage{natbib}
\usepackage[acronym,shortcuts]{glossaries}
\usepackage{listings}
\graphicspath{ {/Figures/} }
\usepackage{float}
\usepackage{color, colortbl}
\usepackage[toc,page]{appendix}
\usepackage[colorinlistoftodos]{todonotes}
\usepackage{xurl}
\usepackage[usenames,dvipsnames]{xcolor}
\usepackage{setspace} 
\usepackage{lipsum}
\usepackage{tikz}
\usepackage{collcell}
\usepackage{arydshln}
\usepackage{mathtools}
\usepackage{color, colortbl}
\usepackage{xcolor}
\DeclareMathOperator{\diag}{diag}

\newcommand{\R}{\mathbb{R}}

\newcommand{\bmu}{\boldsymbol{\mu}}

\newcommand{\bbeta}{\boldsymbol{\beta}}

\newcommand{\bGamma}{\boldsymbol{\Gamma}}

\biboptions{authoryear}

\newacronym{DSM}{DSM}{digital soil mapping}
\newacronym{RF}{RF}{random forest}
\newacronym{MRF}{MRF}{multivariate random forest}
\newacronym{QRF}{QRF}{quantile regression forest}
\newacronym{DirRF}{DirRF}{Dirichlet random forest}

\newacronym{LUCAS}{LUCAS}{Land Use and Coverage Area frame Survey}

\newacronym{CoDA}{CoDA}{compositional data analysis}

\newacronym{alr}{alr}{additive log-ratio}
\newacronym{clr}{clr}{centred log-ratio}
\newacronym{ilr}{ilr}{isometric log-ratio}

\newacronym{mln}{MLN}{multivariate logistic-normal}
\newacronym{cart}{CART}{classification and regression tree}
\newacronym{mmd}{MMD}{maximum mean discrepancy}

\newacronym{ME}{ME}{mean error}
\newacronym{MSE}{MSE}{mean square error}
\newacronym{MAE}{MAE}{mean absolute error}
\newacronym{RMSE}{RMSE}{root mean square error}
\newacronym{MEC}{MEC}{model efficiency coefficient}
\newacronym{CCC}{CCC}{concordance correlation coefficient}

\makenoidxglossaries

\newtheorem{definition}{Definition}
\newtheorem{theorem}{Theorem}
\newtheorem{lemma}{Lemma}

\newtheorem{corollary}{Corollary}

\begin{document}

\begin{frontmatter}

\title{Compositional random forest}
\author[address1,address2]{Khaled Masoumifard\corref{mycorrespondingauthor}}

\cortext[mycorrespondingauthor]{Corresponding author}
\ead{khaledm@sun.ac.za}
\author[address1,address2]{Stephan van der Westhuizen}

\author[address1,address2]{Sugnet Gardner-Lubbe}

\address[address1]{Dept. of Statistics and Actuarial Science, Stellenbosch University, Stellenbosch, South Africa}
\address[address2]{Centre for Multi-dimensional Data Visualisation (MuViSU), Stellenbosch University, Stellenbosch, South Africa}

\begin{abstract}
\noindent
Compositional data have been analysed in the literature from three main
points of view: directly on the simplex, in log-ratio coordinates, and as
directional data on the unit sphere. In this paper, we construct a random
forest framework for compositional responses that accommodates all three.
Random forests are ensembles of decision trees, and the split criterion
determines how each tree grows. We use the log-likelihood gain as the split
criterion and consider distributions defined on the different spaces in
which compositions can be represented. The Dirichlet likelihood has been
studied in the literature, but its covariance is fixed by its mean up to a
single scalar, and its maximum likelihood estimator has no closed form. To
obtain more flexible criteria, we use two well-known transformations. First,
the square-root transformation maps compositions onto the unit sphere, where
existing directional distributions require numerical estimation. We
introduce a tangent-normal (TN) distribution, which shares its parameters
with the Kent distribution, and construct a split criterion whose gain is
computed from additive sufficient statistics: the shape parameters are
maximised out in closed form, up to an exponentially small truncation
constant, so that no parameter is estimated iteratively. Second, the
log-ratio transformation maps compositions to Euclidean space, where the
multivariate logistic-normal distribution has a flexible covariance and a
closed-form maximum log-likelihood, also based on additive sufficient
statistics. Both criteria respond to a shift of the mean, measured relative
to the covariance, and to a change in the covariance itself. In
simulations, the most accurate forest depended on the covariance structure
of the composition. The TN forest was the most accurate in two of the three
designs.
\end{abstract}


\end{frontmatter}


\mathchardef\mhyphen="2D

\section{Introduction} \label{Sec_intro}

\noindent
Compositional data are vectors of non-negative components that sum to a
constant and therefore lie in a simplex space. Multivariate methods developed for
real space are generally inappropriate in this  space
\citep{RefBook_aitchison1986statistical,RefBook_VeraCoDABook2}, and how
compositions should be modelled has been the subject of a long
methodological debate.Three of the perspectives in the compositional data analysis are:
distributions defined directly on the simplex, such as the Dirichlet
distribution; log-ratio transformations, which transfer the simplex to real space
\citep{RefArt_aitchison1982statistical,RefArt_Vera_ILR}; and the square-root
transformation, which transfers compositions onto the positive orthant of the unit
sphere \citep{ScealyWelsh2011,RefArt_ScealyWelsh2014,Yoon2025}.

\noindent
Distributions defined directly on the simplex, such as the Dirichlet, impose
restrictive correlation structures
\citep{RefBook_aitchison1986statistical}. 
\citet{RefArt_aitchison1982statistical,RefBook_aitchison1986statistical}
proposed the log-ratio transformations and the multivariate logistic-normal
(MLN) distribution \citep{RefArt_MLNdist}, a more flexible alternative to the
Dirichlet. Another proposal concerned the directional representation of
compositions. \citet{RefArt_stephens1982use} modelled square-root-transformed
proportions with the von Mises distribution on the sphere, and
\citet{RefArt_watson1989measures} represented compositions as unit vectors
in the positive orthant of the sphere and measured differences between them
by angles rather than by log-ratios. This led to an exchange of comments and
replies in which Aitchison defended the log-ratio approach against these
alternatives
\citep{RefArt_aitchison1990comment,RefArt_watson1990reply,RefArt_aitchison1991delusions,RefArt_watson1991reply}.

\noindent
This debate has influenced the subsequent development of the field.
 Log-ratio methods
became the dominant approach \citep{RefArt_CoDAReview}, and
\citet{RefArt_scealy2014colours} argue that this dominance reflects not only
the advantages of these methods but also the way in which the debate was
conducted: a set of principles, including scale invariance,
subcompositional coherence and permutation invariance, came to serve as a
criterion for whether a method is acceptable for compositional data, so that
methods that did not satisfy these principles were regarded as invalid,
irrespective of their statistical properties or practical usefulness.
They describe this situation as unusual in statistical data analysis.
Methods outside the log-ratio framework have nevertheless continued to be
developed, partly for a practical reason: log-ratio transformations are
undefined when a component is zero, and for highly variable data near the
boundary of the simplex, log-ratio and alternative methods can lead to
substantively different conclusions \citep{RefArt_scealy2014colours}.
In particular, the Dirichlet and directional approaches have received
renewed attention for regression with a compositional response. Examples
include Dirichlet regression models
\citep{RefArt_hijazi2009modelling,RefArt_gueorguieva2008dirichlet,RefTech_maier2014dirichletreg}
and the regression models of \citet{ScealyWelsh2011,Scealy201724}, which fit
the Kent distribution \citep{Kent1982} to square-root-transformed
compositions and handle zero components without replacement. Other
distributions for square-root-transformed compositions include the scaled
von Mises--Fisher distribution \citep{Scealy20191547} and the elliptically
symmetric angular Gaussian (ESAG) distribution \citep{Paine2018,Yu2024301},
and these distributions have been used for regression with directional and
compositional responses
\citep{Scealy20191547,Paine2020153,Yu2025,Schwob20252644}.

\noindent
Random forests \citep{RefArt_BreimanRF}, which are ensembles of decision
trees, have also been adapted to compositional responses. A common approach
is to map the data to real space with a log-ratio transformation and then
apply a random forest. An alternative is to base the split criterion on a
likelihood for the bounded or compositional response itself.
\citet{Beta_RF} developed a random forest based on the beta distribution for
bounded outcomes, and, for compositional responses,
\citet{Khaled_Steohan_2026} proposed the Dirichlet random forest, whose split
criterion is the gain in Dirichlet log-likelihood. However, the Dirichlet
distribution has two limitations. First, its covariance matrix is determined
by its mean up to a single precision parameter, so the Dirichlet criterion
cannot capture changes in the relative variances of the components or in
their dependence structure. Second, its maximum likelihood estimator has no
closed form \citep{RefReport_EstDirDist} and must be computed iteratively for
every candidate child node at every split. A more flexible criterion
therefore requires a distribution with a less constrained covariance
structure whose maximised likelihood is inexpensive to evaluate for every
 candidate. The MLN distribution satisfies both requirements, but
its log-ratio coordinates can become highly variable when a component is
close to zero. Directional distributions for square-root-transformed
compositions behave well near zero, but they also suffer from the lack of a
closed-form maximum likelihood estimator (see, e.g.,
\citealt{Kent1982,ScealyWelsh2011} for the Kent distribution,
\citealt{Scealy20191547} for the scaled von Mises--Fisher distribution, and
\citealt{Paine2018,Yu2024301} for the ESAG distribution). In a forest, this
numerical estimation would have to be repeated for every  candidate  during the split search. Consequently, none of the distributions
reviewed here yields a split criterion that combines a flexible covariance
structure, stable behaviour near zero, and a maximised likelihood that is
inexpensive to evaluate. A practical split criterion therefore requires
either an approximation to the maximum likelihood estimator of an existing
directional distribution or a new distribution that is computationally more
convenient.

\noindent
In this paper, we develop a random forest framework that accommodates all
three perspectives, with three likelihood-based split criteria: the
Dirichlet criterion, the MLN criterion, and a new criterion based on the
tangent-normal (TN) distribution for square-root-transformed compositions.
The maximised MLN log-likelihood has a closed form. For the TN criterion, we
define a tangent-normal distribution on the sphere and show that, given the
moment estimate of its mean direction, its maximised log-likelihood admits a
closed-form approximation. We then show that both the MLN and TN criteria
reward shifts in the mean, measured relative to the covariance, as well as
changes in the covariance itself. Unlike the Dirichlet criterion, these two
criteria can therefore respond to covariates that change the relative
variances or the dependence structure of the composition. Finally, we
compare the performance of the three forests through experiments on
simulated datasets.

\section{Compositional data and random forests} \label{Sec_Method}
\subsection{Compositional data and transformations}\label{sec:coda}
\noindent
Consider a random vector $\mathbf{Y} = (Y_1, \ldots, Y_k) \in \R_+^k$ with $k$ the number of components. The closure of $\mathbf{Y}$ to a constant $c$ is defined as
\begin{align}
\mathcal{C}_c(\mathbf{Y}) &= \left(\frac{c Y_1}{\sum_{j=1}^k Y_j}, \ldots, \frac{c Y_k}{\sum_{j=1}^k Y_j}\right), \notag \\
&:= (Z_1,\cdots,Z_k). \label{eq:Z_vect}
\end{align}
Therefore, the observation $\mathbf{z} = (z_1, \ldots, z_k)$ belongs to the $k$-dimensional simplex,

\begin{equation}
   \mathbf{S}^k = \Bigl\{(z_1, \ldots, z_k) : z_j>0,\ \sum_{j=1}^k z_j = c\Bigr\}.
\label{eq:simplex}
\end{equation}

\noindent
Furthermore, when the constant $c$ carries no relevant information, it is standard practice to set $c=1$ without loss of generality. Data belonging to this space are referred to as compositional data. Because of the limitations associated with the simplex space, it is common to transform data from $\mathbf{S}^k$ to a more flexible space using a suitable transformation before modelling.
The following transformations are well known and widely used. The
\emph{additive log-ratio} (ALR) transformation transforms $\mathbf{S}^k$ to
$\R^{k-1}$ by taking log-ratios relative to a fixed reference component,
such as $z_k$,
\begin{equation}
y_j = \ln\!\left(\frac{z_j}{z_k}\right), \qquad j = 1,\ldots,k-1.
\end{equation}
The \emph{centred log-ratio} (CLR) transformation instead normalises each
component by the geometric mean of the composition,
\begin{equation}
y_j = \ln\!\left(\frac{z_j}{g(\mathbf{z})}\right), \qquad
g(\mathbf{z}) = \Big(\prod_{\ell=1}^k z_\ell\Big)^{1/k}, \qquad j = 1,\ldots,k,
\end{equation}
producing a symmetric, reference-free representation at the cost of
mapping into a $(k-1)$-dimensional subspace of $\R^k$ subject to the
constraint $\sum_j y_j = 0$. Consequently, the induced covariance matrix
is singular. The \emph{isometric log-ratio} (ILR) transformation resolves
this issue by projecting the CLR coordinates onto an orthonormal basis of
the $(k-1)$-dimensional hyperplane they span. We do not discuss the technical details further here and refer
the reader to \citet{RefBook_VeraCoDABook2} for details.

\noindent
A non-log-ratio alternative for compositional data is the
\emph{square-root transformation}, applied
componentwise, which maps $\mathbf{z} \in \mathbf{S}^k$ to
$\mathbf{u}=\sqrt{\mathbf{z}}$ on the $(k-1)$-dimensional hypersphere
\[
   S^{k-1} \;=\; \bigl\{\mathbf{u} \in \R^k : \|\mathbf{u}\| = 1\bigr\}.
\]
This allows the use of distributions whose natural support is $S^{k-1}$,
and in particular of methods from the directional-data literature. Under
this transformation, compositions with structural zeros are admissible
without the replacement or perturbation that any log-ratio transformation
requires. Following common practice, we take the non-negative root
throughout, so that compositions are mapped into the positive orthant
\[
   \mathcal{O} \;=\; \bigl\{\mathbf{u} \in S^{k-1} : u_j \ge 0 \ \text{for all } j\bigr\}
\]
rather than onto the whole of $S^{k-1}$.



\subsection{Random forests with likelihood-based splits}\label{sec:rf}

Let $\mathbf{x}\in\R^{p}$ be a vector of covariates and
$\mathbf{Z}\in\mathbf{S}^{k}$ the compositional response, and let
$(\mathbf{x}^{(t)},\mathbf{z}^{(t)})$, $t=1,\ldots,n$, be the training data.

\noindent
A decision tree is grown by recursive binary partitioning. Starting from a
root node that contains all training observations, each node $\Omega$ is
split by a covariate $j\in\{1,\ldots,p\}$ and a threshold $\theta$ into the
child nodes
\[
   L(\theta,j)=\{t\in\Omega: x_j^{(t)}\le\theta\},
   \qquad
   R(\theta,j)=\{t\in\Omega: x_j^{(t)}>\theta\},
\]
and then the child nodes are split in the same way. Nodes that the stopping
rule does not allow to be split further are the leaves, or terminal nodes. A
random forest \citep{RefArt_BreimanRF} is an ensemble of $B$ such trees,
each grown on a resample of the training data, or on all of it, with a
random subset of $m_{\mathrm{try}}$ candidate covariates at each split.

\noindent The split criterion determines which candidate split $(\theta,j)$ is selected and may vary depending on the purpose of the analysis and the geometry of the data. For compositional data, following \citet{Beta_RF,Khaled_Steohan_2026,Stephan_Khaled_2026}, we use likelihood-based criteria. Let $\hat\ell_F(A)$ denote the maximum log-likelihood under the distributional model $F$ fitted to the observations in a node $A$. The selected split is
\begin{equation}\label{eq:general-split}
   (\theta^{*},j^{*})
   =\arg\max_{\theta,j}\,\Delta\ell_F(\theta,j),
   \qquad
   \Delta\ell_F(\theta,j)
   =\hat\ell_F\bigl(L(\theta,j)\bigr)+\hat\ell_F\bigl(R(\theta,j)\bigr)
    -\hat\ell_F(\Omega).
\end{equation}
Here, $\Omega$ denotes the parent node, while $L(\theta,j)$ and $R(\theta,j)$ denote the two child nodes induced by the candidate split. The quantity $\Delta\ell_F(\theta,j)$ measures the improvement in maximum log-likelihood obtained by splitting the parent node into two children. A split is rewarded when two distributions describe the data of the
node better than one, through a difference in location (the mean,
or, for square-root coordinates, the mean direction) or in spread and
dependence, to the extent that $F$ can represent them. The choice of $F$,
and how it determines what the criterion rewards, is the subject of
Section~\ref{sec:nodes}.

To predict at a new point $\mathbf{x}$, each tree routes $\mathbf{x}$ to a
leaf $L_b(\mathbf{x})$, $b=1,\ldots,B$, and the forest assigns to training
observation $t$ the weight \citep{Lin-RandomForestsAdaptive-2006}
\begin{equation}\label{eq:forest-weights}
   w_t(\mathbf{x})
   =\frac{1}{B}\sum_{b=1}^{B}
    \frac{\mathbf{1}\{\mathbf{x}^{(t)}\in L_b(\mathbf{x})\}}{|L_b(\mathbf{x})|},
   \qquad \sum_{t=1}^{n}w_t(\mathbf{x})=1 .
\end{equation}
As in the distributional random forest of \citet{RefArt_drf}, which uses
weights of the same form, the conditional distribution of $\mathbf{Z}$ given
$\mathbf{x}$ is estimated by the weighted empirical distribution
\begin{equation}\label{eq:empirical-conditional}
   \hat P(\mathbf{z}\mid\mathbf{x})
   =\sum_{t=1}^{n}w_t(\mathbf{x})\,  {1}\{\mathbf{z}^{(t)}\le\mathbf{z}\}.
\end{equation}
Any functional of the conditional
distribution follows from \eqref{eq:empirical-conditional}; in particular,
the point prediction is its mean,
$\hat{\mathbf{z}}(\mathbf{x})=\sum_t w_t(\mathbf{x})\,\mathbf{z}^{(t)}$, which
lies in $\mathbf{S}^{k}$. 

\section{Distributions and split criteria}\label{sec:nodes}

\subsection{Dirichlet}\label{sec:dir}

The Dirichlet distribution is a natural choice for modelling compositional
data. For $\mathbf{Z}\sim\operatorname{Dir}(\boldsymbol{\alpha})$, with
$\boldsymbol{\alpha}=(\alpha_1,\ldots,\alpha_k)$ and $\alpha_j>0$, the
density on $\mathbf{S}^k$ is
\begin{equation}\label{eq:dir-density}
   f(\mathbf{z};\boldsymbol{\alpha})
   =\frac{\Gamma(\alpha_0)}{\prod_{j=1}^{k}\Gamma(\alpha_j)}
    \prod_{j=1}^{k}z_j^{\alpha_j-1},
   \qquad
   \alpha_0=\sum_{j=1}^{k}\alpha_j .
\end{equation}
Writing $\boldsymbol{\alpha}=\alpha_0\mathbf{m}$, with mean
$\mathbf{m}\in\mathbf{S}^k$ and precision $\alpha_0$, the first two moments
are 
\begin{equation}\label{eq:dir-moments}
   E(\mathbf{Z})=\mathbf{m},
   \qquad
   \operatorname{Cov}(\mathbf{Z})
   =\frac{\operatorname{diag}(\mathbf{m})-\mathbf{m}\mathbf{m}^{\top}}{\alpha_0+1}.
\end{equation}

\noindent
The Dirichlet random forest of \citet{Khaled_Steohan_2026} uses the Dirichlet
log-likelihood as split criterion. The maximum
likelihood estimators of $\boldsymbol{\alpha}$ have no closed form and must be
computed iteratively \citep{RefReport_EstDirDist}, for all candidate splits. This can be very computationally expensive and therefore we use 
the method of moments instead \citep{Stephan_Khaled_2026}.

\noindent
By \eqref{eq:dir-moments}, 
the covariance matrix is determined by the mean and the single scalar
$1/(\alpha_0+1)$, and all correlations are negative. Moreover, by the delta
method, for large $\alpha_0$,
\begin{equation}\label{eq:dir-isotropic}
   \operatorname{Cov}\bigl(\sqrt{\mathbf{Z}}\bigr)
   \approx\frac{\mathbf{I}_k-\sqrt{\mathbf{m}}\,\sqrt{\mathbf{m}}^{\top}}{4(\alpha_0+1)},
\end{equation}
so that, in square-root coordinates, a Dirichlet distribution has
approximately the same variance in every direction of the tangent plane at
$\sqrt{\mathbf{m}}$. This rigid covariance structure is a well-known
limitation of the Dirichlet distribution
\citep{RefBook_aitchison1986statistical}. Data whose covariance does not
follow this structure need a more flexible distribution, whose covariance
is not tied to its mean.

\subsection{Tangent-normal}\label{sec:kent}

\noindent As noted in Section~\ref{Sec_Method}, the componentwise square root
$\mathbf{u}=\sqrt{\mathbf{z}}$ maps compositions one-to-one into the
positive orthant $\mathcal{O}$ of the unit sphere $S^{k-1}$.
A natural choice for $\mathbf{u}$
is the Kent distribution \citep{Kent1982}, in the general-dimensional form of
\citet{ScealyWelsh2011},
\begin{equation}\label{eq:kent-density}
   f(\mathbf{u})
   =
   c(\kappa,\bbeta)^{-1}
   \exp\!\Bigl[
      \kappa\,\bmu^{\top}\mathbf{u}
      + \sum_{m=2}^{k-1}\beta_m\bigl(\boldsymbol{\gamma}_m^{\top}\mathbf{u}\bigr)^{2}
      - \Bigl(\sum_{m=2}^{k-1}\beta_m\Bigr)\bigl(\boldsymbol{\gamma}_k^{\top}\mathbf{u}\bigr)^{2}
   \Bigr],
   \quad \mathbf{u} \in S^{k-1},
\end{equation}
where $\bmu\in S^{k-1}$ is the mean direction and $\kappa>0$ the
concentration. The elements of
$(\boldsymbol{\gamma}_2,\ldots,\boldsymbol{\gamma}_k)$ are $k-1$
orthonormal vectors orthogonal to $\bmu$ and determine the directions along which the spread varies, while the shape parameters
$\bbeta=(\beta_2,\ldots,\beta_{k-1})^{\top}$ control how much it differs.
With
\begin{equation}\label{eq:beta-k}
   \beta_k:=-\sum_{m=2}^{k-1}\beta_m,
\end{equation}
the shape parameters sum to zero, and they should satisfy
\citep[eq.~(3)]{ScealyWelsh2011}
\begin{equation}\label{eq:kent-constraint}
   \tfrac{\kappa}{2} > \beta_2 \ge \beta_3 \ge \cdots \ge \beta_{k-1} \ge \beta_k .
\end{equation}
The special case $\bbeta=\mathbf{0}$ is the von Mises--Fisher distribution,
whose spread is the same in every direction.

Working with the Kent distribution raises two challenges. First, the
density \eqref{eq:kent-density} is positive on the whole sphere, whereas
compositions lie in $\mathcal{O}$, so a Kent model gives positive
probability to points with negative coordinates, which no composition can
produce. Second, the split criterion \eqref{eq:general-split} needs the
maximum log-likelihood of every candidate child, and for the Kent
distribution this is computationally expensive for
every candidate child of every split. \citet[Theorem~3]{ScealyWelsh2011}
show that, as the concentration grows, the Kent distribution becomes normal
in the tangent coordinates. Motivated by this result, we
define a distribution that is exactly normal in these coordinates. There are other
Kent-like distributions with tractable likelihoods, for example the
scaled von Mises--Fisher distribution \citep{Scealy20191547} and the
elliptically symmetric angular Gaussian distribution
\citep{Paine2018,Yu2024301}, but their shape parameters must be estimated by
numerical optimisation, which a split criterion would repeat for every
candidate child.

\subsubsection{The tangent-normal distribution}\label{sec:tn-def}
 
\noindent
Define the open unit ball in $\R^{k-1}$ and the open hemisphere centred at
$\bmu\in S^{k-1}$ as
\[
   \mathbb{B}^{k-1}=\{\mathbf{v}\in\R^{k-1}:\|\mathbf{v}\|<1\},
   \qquad
   H_{\bmu}=\{\mathbf{u}\in S^{k-1}:\bmu^{\top}\mathbf{u}>0\}.
\]
Let $\mathbf{K}$ be a $k\times(k-1)$ matrix with
$\mathbf{K}^{\top}\mathbf{K}=\mathbf{I}_{k-1}$ whose columns span the
tangent space
\[
   T_{\bmu}=\{\mathbf{x}\in\R^{k}:\bmu^{\top}\mathbf{x}=0\}.
\]
We now explore how data can be mapped from one space to another.
\begin{lemma}\label{lem:tn-chart}
The matrix $\bGamma=(\bmu,\mathbf{K})$ is orthogonal, and 
\begin{equation}\label{eq:kent-return}
   \psi(\mathbf{v})
   =\bigl(1-\|\mathbf{v}\|^{2}\bigr)^{1/2}\,\bmu+\mathbf{K}\mathbf{v},
   \qquad \mathbf{v}\in\mathbb{B}^{k-1},
\end{equation}
is a bijection from $\mathbb{B}^{k-1}$ onto $H_{\bmu}$, with inverse
$\psi^{-1}(\mathbf{u})=\mathbf{K}^{\top}\mathbf{u}$. Moreover, $\psi$ is
continuously differentiable and
$\bmu^{\top}\psi(\mathbf{v})=(1-\|\mathbf{v}\|^{2})^{1/2}$.
\end{lemma}

\begin{proof}
Since the columns of $\mathbf{K}$ lie in $T_{\bmu}$,
$\mathbf{K}^{\top}\bmu=\mathbf{0}$; together with $\bmu^{\top}\bmu=1$ and
$\mathbf{K}^{\top}\mathbf{K}=\mathbf{I}_{k-1}$, this gives
$\bGamma^{\top}\bGamma=\mathbf{I}_k$, and, as $\bGamma$ is square, also
$\bGamma\bGamma^{\top}=\bmu\bmu^{\top}+\mathbf{K}\mathbf{K}^{\top}=\mathbf{I}_k$.

\noindent
\emph{From $\mathbb{B}^{k-1}$ to $H_{\bmu}$.} Let
$\mathbf{v}\in\mathbb{B}^{k-1}$ and $\mathbf{u}=\psi(\mathbf{v})$. Using
$\mathbf{K}^{\top}\bmu=\mathbf{0}$ and
$\mathbf{K}^{\top}\mathbf{K}=\mathbf{I}_{k-1}$,
\[
   \|\mathbf{u}\|^{2}
   =\bigl(1-\|\mathbf{v}\|^{2}\bigr)
    +2\bigl(1-\|\mathbf{v}\|^{2}\bigr)^{1/2}\bmu^{\top}\mathbf{K}\mathbf{v}
    +\mathbf{v}^{\top}\mathbf{K}^{\top}\mathbf{K}\mathbf{v}
   =1,
   \qquad
   \bmu^{\top}\mathbf{u}=\bigl(1-\|\mathbf{v}\|^{2}\bigr)^{1/2}>0,
\]
so $\mathbf{u}\in H_{\bmu}$. Moreover,
$\mathbf{K}^{\top}\mathbf{u}
=(1-\|\mathbf{v}\|^{2})^{1/2}\mathbf{K}^{\top}\bmu+\mathbf{K}^{\top}\mathbf{K}\mathbf{v}
=\mathbf{v}$, that is, $\mathbf{K}^{\top}\psi(\mathbf{v})=\mathbf{v}$ for
every $\mathbf{v}\in\mathbb{B}^{k-1}$, so $\psi$ is one-to-one.

\noindent
\emph{From $H_{\bmu}$ to $\mathbb{B}^{k-1}$.} Let $\mathbf{u}\in H_{\bmu}$
and $\mathbf{v}=\mathbf{K}^{\top}\mathbf{u}$. Since $\bGamma$ is orthogonal,
\[
   1=\|\mathbf{u}\|^{2}=\|\bGamma^{\top}\mathbf{u}\|^{2}
   =(\bmu^{\top}\mathbf{u})^{2}+\|\mathbf{v}\|^{2}.
\]
As $\bmu^{\top}\mathbf{u}>0$, it follows that $\|\mathbf{v}\|<1$, so
$\mathbf{v}\in\mathbb{B}^{k-1}$, and
$\bmu^{\top}\mathbf{u}=(1-\|\mathbf{v}\|^{2})^{1/2}$, taking the positive
root. Then $\bGamma\bGamma^{\top}=\mathbf{I}_k$ gives
\[
   \mathbf{u}
   =\bmu\bmu^{\top}\mathbf{u}+\mathbf{K}\mathbf{K}^{\top}\mathbf{u}
   =\bigl(1-\|\mathbf{v}\|^{2}\bigr)^{1/2}\bmu+\mathbf{K}\mathbf{v}
   =\psi(\mathbf{v}),
\]
so $\psi$ is onto $H_{\bmu}$, and $\psi(\mathbf{K}^{\top}\mathbf{u})=\mathbf{u}$.

\noindent
Finally, $\psi$ is continuously differentiable on $\mathbb{B}^{k-1}$,
because $1-\|\mathbf{v}\|^{2}>0$.
\end{proof}

\noindent
The map $\psi$ has a simple geometric interpretation, which also explains
the roles of $\bmu$, $\mathbf{K}$ and $\mathbf{v}$. Every
$\mathbf{u}\in H_{\bmu}$ can be written as
$\mathbf{u}=(\cos\theta)\,\bmu+(\sin\theta)\,\mathbf{e}$, where
$\theta\in[0,\pi/2)$ is the angle between $\mathbf{u}$ and $\bmu$ and
$\mathbf{e}\in T_{\bmu}$ is a unit vector. Then
$\bmu^{\top}\mathbf{u}=\cos\theta$ and
$\mathbf{v}=\mathbf{K}^{\top}\mathbf{u}=(\sin\theta)\,\mathbf{K}^{\top}\mathbf{e}$,
so $\mathbf{v}$ holds the coordinates, in the basis $\mathbf{K}$, of the
orthogonal projection of $\mathbf{u}$ onto the tangent plane $T_{\bmu}$; we
call them the \emph{tangent coordinates} of $\mathbf{u}$. 


\noindent 
Now, let
\[
   \mathbf{V}_0\sim N_{k-1}(\mathbf{0},\Lambda),
   \qquad
   \Lambda=\diag(\lambda_2,\ldots,\lambda_k),
   \qquad
   \lambda_2\ge\cdots\ge\lambda_k>0 ,
\]
and
$
\mathbf{V}\overset{d}{=}(\mathbf{V}_0\mid \|\mathbf{V}_0\|<1)
$, where \(\overset{d}{=}\) denotes equality in distribution, and set
\[
   \mathbf{U}=\bigl(1-\|\mathbf{V}\|^{2}\bigr)^{1/2}\bmu+\mathbf{K}\mathbf{V}.
\]
By definition and Lemma~\ref{lem:tn-chart}, $\mathbf{V}\in\mathbb{B}^{k-1}$ and $\mathbf{U}\in H_{\bmu}$
almost surely.
 
\begin{definition}\label{def:tn}
We refer to the distribution of
$\mathbf{U}$ as the {tangent-normal} distribution with mean direction
$\bmu$, axes $\mathbf{K}$ and tangent variances $\Lambda$, written
$\mathbf{U}\sim\mathrm{TN}(\bmu,\mathbf{K},\Lambda)$.
\end{definition}
 
\begin{lemma}\label{lem:tn-density}
Let $\pi_{\Lambda}=P(\|\mathbf{V}_0\|<1)$ and let $\varphi_{\Lambda}$ be the
density of $N_{k-1}(\mathbf{0},\Lambda)$. The density of
$\mathbf{U}\sim\mathrm{TN}(\bmu,\mathbf{K},\Lambda)$ with respect to the
surface measure $\sigma$ on $S^{k-1}$ is
\begin{equation}\label{eq:kent-working-density}
   g(\mathbf{u})
   =\frac{\varphi_{\Lambda}(\mathbf{K}^{\top}\mathbf{u})\;\bmu^{\top}\mathbf{u}}{\pi_{\Lambda}},
   \qquad \mathbf{u}\in H_{\bmu},
\end{equation}
and $g(\mathbf{u})=0$ outside $H_{\bmu}$; that is,
$P(\mathbf{U}\in E)=\int_E g\,d\sigma$ for every Borel set
$E\subset S^{k-1}$.
\end{lemma}

\begin{proof}
By Lemma~\ref{lem:tn-chart}, $\mathbf{U}=\psi(\mathbf{V})$, where $\psi$ is
a continuously differentiable bijection from $\mathbb{B}^{k-1}$ onto
$H_{\bmu}$ with inverse $\mathbf{u}\mapsto\mathbf{K}^{\top}\mathbf{u}$, and
$\mathbf{V}$ has density $\varphi_{\Lambda}(\mathbf{v})/\pi_{\Lambda}$ on
$\mathbb{B}^{k-1}$. Since $\mathbf{K}^{\top}\mathbf{K}=\mathbf{I}_{k-1}$ and
$\mathbf{K}^{\top}\bmu=\mathbf{0}$, the derivative
$D\psi(\mathbf{v})=\mathbf{K}-\bmu\mathbf{v}^{\top}/(1-\|\mathbf{v}\|^{2})^{1/2}$
satisfies, by the matrix determinant lemma,
\[
   \det\bigl(D\psi^{\top}D\psi\bigr)
   =\det\Bigl(\mathbf{I}_{k-1}+\frac{\mathbf{v}\mathbf{v}^{\top}}{1-\|\mathbf{v}\|^{2}}\Bigr)
   =\frac{1}{1-\|\mathbf{v}\|^{2}}.
\]
By the change-of-variables formula for parametrised surfaces
\citep[Sec.~3.3]{EvansGariepy2015},
$\sigma(d\mathbf{u})=\det(D\psi^{\top}D\psi)^{1/2}\,d\mathbf{v}$, that is,
\begin{equation}\label{eq:tn-jacobian}
   d\mathbf{v}=(\bmu^{\top}\mathbf{u})\,\sigma(d\mathbf{u}),
\end{equation}
because $\bmu^{\top}\mathbf{u}=(1-\|\mathbf{v}\|^{2})^{1/2}$. Hence $\mathbf{U}$
has density
$\pi_{\Lambda}^{-1}\varphi_{\Lambda}(\mathbf{K}^{\top}\mathbf{u})\,\bmu^{\top}\mathbf{u}$
with respect to $\sigma$ on $H_{\bmu}$, and zero outside $H_{\bmu}$.
\end{proof}

\noindent
The factor $\bmu^{\top}\mathbf{u}$ in \eqref{eq:kent-working-density} is
the Jacobian \eqref{eq:tn-jacobian}: it turns the density of the tangent
coordinates, with respect to Lebesgue measure on $T_{\bmu}$, into a density
on the sphere. 

\subsubsection{Relationship with the Kent and other spherical distributions}
\label{sec:tn-kent}
\noindent
The tangent-normal distribution has the same location and orientation
parameters as the Kent distribution, and if we set
\begin{equation}\label{eq:lambda-def}
   \lambda_m=\bigl(\kappa-2\beta_m\bigr)^{-1},
   \qquad m=2,\ldots,k,
\end{equation}
then the tangent variances correspond one-to-one to the concentration and
shape parameters of the Kent distribution. Because $\sum_{m}\beta_m=0$,
$\kappa=(k-1)^{-1}\sum_{m}\lambda_m^{-1}$ and
$\beta_m=\tfrac12(\kappa-\lambda_m^{-1})$, and \eqref{eq:kent-constraint}
holds exactly when $\lambda_2\ge\cdots\ge\lambda_k>0$. The two distributions
therefore have the same parameter space, with $\tfrac12k(k+1)-1$ free
parameters.

\noindent
The two distributions also share the structure of their moments: $E(\mathbf{V})=\mathbf{0}$,
$E\{(1-\|\mathbf{V}\|^{2})^{1/2}\mathbf{V}\}=\mathbf{0}$ and $E(V_mV_l)=0$
for $m\ne l$, and by \eqref{eq:kent-return}
\begin{equation}\label{eq:tn-moments}
   E(\mathbf{U})=E\bigl\{(1-\|\mathbf{V}\|^{2})^{1/2}\bigr\}\,\bmu,
   \qquad
   E(\mathbf{U}\mathbf{U}^{\top})
   =\bGamma\diag(\tau_1,\ldots,\tau_k)\,\bGamma^{\top},
\end{equation}
with $\tau_m=E(V_m^{2})$ for $m\ge2$ and $\tau_1=1-\sum_{m=2}^{k}\tau_m$; the
Kent distribution has the same structure \citep[eq.~(9)]{ScealyWelsh2011}.

\noindent
The two distributions have the same normal limit in the tangent
coordinates. \citet[Theorem~3]{ScealyWelsh2011} show that if
$\kappa\to\infty$ with $\beta_m/\kappa\to d_m$, where
$\tfrac12>d_2\ge\cdots\ge d_k$, then under the Kent distribution the
$\kappa^{1/2}v_m$ are asymptotically independent $N\{0,(1-2d_m)^{-1}\}$.
The tangent-normal distribution with $\lambda_m$ given by
\eqref{eq:lambda-def} has the same limit.
By Markov's inequality,
\begin{equation}\label{eq:tn-chernoff}
\begin{aligned}
1-\pi_\Lambda
&=P\!\left(\|V_0\|^2\ge 1\right)
\le \frac{\mathbb{E}\!\left[\exp\!\left(\frac{\|V_0\|^2}{4\lambda_2}\right)\right]}
{\exp\!\left(\frac{1}{4\lambda_2}\right)},
\end{aligned}
\end{equation}
and
\[
\mathbb{E}\!\left[\exp\!\left(\frac{\|V_0\|^2}{4\lambda_2}\right)\right]
=\prod_{m=2}^{k}\left(1-\frac{\lambda_m}{2\lambda_2}\right)^{-1/2}
\leq \prod_{m=2}^{k}2^{1/2}
=2^{(k-1)/2}.
\]
Hence $1-\pi_\Lambda \le 2^{\frac{k-1}{2}}\exp\!\left(-\frac{1}{4\lambda_2}\right)$.
Under \eqref{eq:lambda-def} and the conditions of \citet[Theorem~3]{ScealyWelsh2011} stated above,
\[
\lambda_2=\kappa^{-1}\left(1-\frac{2\beta_2}{\kappa}\right)^{-1}\longrightarrow 0,
\]
so that $1-\pi_\Lambda\to0$ exponentially fast.

\noindent
It is worth noting that the two distributions differ in several respects.
The Kent
distribution is defined on the whole sphere, whereas the TN distribution is
supported only on the open hemisphere $H_{\bmu}$, exactly half of the
sphere. The TN distribution has a computational advantage,
for example, $\pi_\Lambda$ is the
distribution function of a weighted sum of $\chi^2_1$ variables at one,
which is computed accurately by the series of \citet{Ruben1962}, whereas
the normalising constant of the Kent distribution is a $(k-1)$-dimensional
integral with no closed form.

\noindent
The omitted TN constant is also much smaller than the corresponding Kent constant (Appendix \ref{app:constants}), so dropping it from the split criterion costs less.

\subsubsection{Split criterion}\label{sec:kent-computation}
 
\noindent 
We assume that, in a node $A$, the observations $\mathbf{u}^{(t)}$,
$t\in A$, are independent with distribution
$\mathrm{TN}(\bmu,\mathbf{K},\Lambda)$. By \eqref{eq:kent-working-density},
the log-likelihood of the node is
\begin{equation}\label{eq:kent-working-loglik}
   \ell(A)+J(A;\bmu)-n_A\ln\pi_{\Lambda},
   \qquad
   \ell(A)=\sum_{t\in A}\ln\varphi_{\Lambda}\bigl(\mathbf{K}^{\top}\mathbf{u}^{(t)}\bigr),
   \qquad
   J(A;\bmu)=\sum_{t\in A}\ln\bigl(\bmu^{\top}\mathbf{u}^{(t)}\bigr).
\end{equation}
Since the term \(n_A\ln\pi_{\Lambda}\) has no closed-form expression and is exponentially small by \eqref{eq:tn-chernoff}, we neglect it in the node-splitting criterion for computational efficiency.
The Jacobian term, by contrast, plays an important role: \(\ell(A)\) is a log-density with respect to Lebesgue measure on \(T_{\bmu}\), which varies with \(\bmu\) and hence from node to node, and adding \(J(A;\bmu)\) expresses the log-density with respect to the common reference measure \(\sigma\) across all nodes.

\noindent 
We now construct the split criterion. Define
\begin{equation}\label{eq:kent-suffstat}
   n_A=|A|,
   \qquad
   \mathbf{s}_1(A)=\sum_{t\in A}\mathbf{u}^{(t)},
   \qquad
   S_2(A)=\sum_{t\in A}\mathbf{u}^{(t)}\mathbf{u}^{(t)\top},
\end{equation}
and let \(\hat M(A)=S_2(A)/n_A\). By \eqref{eq:tn-moments}, \(\bmu\) is the direction of \(E(\mathbf{U})\). Following \citet{ScealyWelsh2011}, we therefore estimate \(\bmu\) using the method of moments, yielding the normalised resultant
\begin{equation}\label{eq:tn-muhat}
\hat{\bmu}(A)=\mathbf{s}_1(A)/\lVert\mathbf{s}_1(A)\rVert .
\end{equation}
This estimator is not the maximum likelihood estimator, which does not admit a closed-form expression for \(\bmu\).
 
\begin{theorem}
\label{thm:kent-llhat}
Let $A$ be a node with $\hat M(A)$ positive definite, and fix the
mean direction at $\hat{\bmu}=\hat{\bmu}(A)$. For all orientations $\mathbf{K}$ with columns forming an orthonormal basis of the tangent plane $T_{\hat{\bmu}}$, and all $\Lambda=\diag(\lambda_2,\ldots,\lambda_k)$ with $\lambda_m>0$, the maximum of $\ell(A)$ is
\begin{equation}\label{eq:kent-llhat}
   \hat\ell_{\mathrm{TN}}(A)
   =
   -\frac{n_A(k-1)}{2}\bigl(1+\ln 2\pi\bigr)
   -\frac{n_A}{2}\ln\Bigl\{\bigl|\hat M(A)\bigr|\,
     \hat{\bmu}^{\top}\hat M(A)^{-1}\hat{\bmu}\Bigr\}.
\end{equation}
\end{theorem}
 
\begin{proof}
Let $\mathbf{B}$ be a $k\times(k-1)$ matrix whose columns are
an orthonormal basis of $T_{\hat{\bmu}}$.
The vector $\mathbf{w}^{(t)}=\mathbf{B}^{\top}\mathbf{u}^{(t)}\in\R^{k-1}$
holds the coordinates of the orthogonal projection of $\mathbf{u}^{(t)}$
onto $T_{\hat{\bmu}}$. Every orientation whose columns are an orthonormal
basis of $T_{\hat{\bmu}}$ can be written as $\mathbf{K}=\mathbf{B}\mathbf{Q}$,
with $\mathbf{Q}=\mathbf{B}^{\top}\mathbf{K}$ orthogonal, so that
$\mathbf{K}^{\top}\mathbf{u}^{(t)}=\mathbf{Q}^{\top}\mathbf{w}^{(t)}$. Define
$\boldsymbol{\Sigma}=\mathbf{Q}\Lambda\mathbf{Q}^{\top}$. Then
$|\boldsymbol{\Sigma}|=|\Lambda|$ and
$(\mathbf{Q}^{\top}\mathbf{w})^{\top}\Lambda^{-1}(\mathbf{Q}^{\top}\mathbf{w})
=\mathbf{w}^{\top}\boldsymbol{\Sigma}^{-1}\mathbf{w}$, so
$\varphi_{\Lambda}(\mathbf{Q}^{\top}\mathbf{w})=\varphi_{\boldsymbol{\Sigma}}(\mathbf{w})$
and
\begin{equation}\label{eq:kent-normal-loglik}
   \ell(A)
   =\sum_{t\in A}\ln\varphi_{\boldsymbol{\Sigma}}\bigl(\mathbf{w}^{(t)}\bigr)
   =-\frac{n_A}{2}\Bigl\{(k-1)\ln2\pi+\ln|\boldsymbol{\Sigma}|
     +\operatorname{tr}\bigl(\boldsymbol{\Sigma}^{-1}S_T(A)\bigr)\Bigr\},
\end{equation}
where $S_T(A)=n_A^{-1}\sum_{t\in A}\mathbf{w}^{(t)}\mathbf{w}^{(t)\top}
=\mathbf{B}^{\top}\hat M(A)\,\mathbf{B}$ is the second moment of the
projected observations, positive definite because $\hat M(A)$ is. Thus
$\ell(A)$ is the log-likelihood of the model in which the projected
observations $\mathbf{w}^{(t)}$ are independent
$N_{k-1}(\mathbf{0},\boldsymbol{\Sigma})$. As $\mathbf{Q}$ and $\Lambda$ vary,
$\boldsymbol{\Sigma}$ ranges over all positive definite matrices. Therefore
$\ell(A)$ is maximised at $\hat{\boldsymbol{\Sigma}}=S_T(A)$, where the trace in \eqref{eq:kent-normal-loglik}
equals $k-1$, so
\begin{equation}\label{eq:kent-llhat-ST}
   \max_{\mathbf{K},\Lambda}\ell(A)
   =-\frac{n_A(k-1)}{2}\bigl(1+\ln2\pi\bigr)-\frac{n_A}{2}\ln\bigl|S_T(A)\bigr|.
\end{equation}
It therefore remains to show that
\[
   |S_T(A)|=|\hat M(A)|\,\hat{\bmu}^{\top}\hat M(A)^{-1}\hat{\bmu}.
\]
Consider the Householder reflector \citep{GolubVanLoan2013}
\begin{equation}\label{eq:kent-householder}
   \mathbf{H}=\mathbf{I}-\frac{2\,\mathbf{h}\mathbf{h}^{\top}}{\mathbf{h}^{\top}\mathbf{h}},
   \qquad \mathbf{h}=\mathbf{e}_1-\hat{\bmu},
\end{equation}
which is symmetric and orthogonal with $\mathbf{H}\mathbf{e}_1=\hat{\bmu}$.
Each of the last $k-1$ columns
$\mathbf{H}\mathbf{e}_j$ is therefore a unit vector orthogonal to $\hat{\bmu}$, that
is, they form an orthonormal basis of
$T_{\hat{\bmu}}$. Since $\mathbf{B}$ was arbitrary, we set
$\mathbf{H}=(\hat{\bmu},\mathbf{B})$. 
Then, 
\begin{equation}\label{eq:kent-block}
   \mathbf{H}\hat M\mathbf{H}
=\begin{pmatrix}\hat{\bmu}^{\top}\\ \mathbf{B}^{\top}\end{pmatrix}
    \hat M\,\bigl(\hat{\bmu}\;\;\mathbf{B}\bigr)
   =\begin{pmatrix}
      \hat{\bmu}^{\top}\hat M\hat{\bmu} & \hat{\bmu}^{\top}\hat M\mathbf{B}\\
      \mathbf{B}^{\top}\hat M\hat{\bmu} & \mathbf{B}^{\top}\hat M\mathbf{B}
    \end{pmatrix}
   =\begin{pmatrix} a & \mathbf{c}^{\top}\\ \mathbf{c} & S_T\end{pmatrix},
\end{equation}
with $a=\hat{\bmu}^{\top}\hat M\hat{\bmu}$ and
$\mathbf{c}=\mathbf{B}^{\top}\hat M\hat{\bmu}$. Since $\mathbf{H}$ is
symmetric and orthogonal, we have
\begin{equation}\label{eq:kent-two-quantities}
   |\mathbf{H}\hat M\mathbf{H}|=|\hat M|,
   \qquad
   \mathbf{e}_1^{\top}(\mathbf{H}\hat M\mathbf{H})^{-1}\mathbf{e}_1
   =(\mathbf{H}\mathbf{e}_1)^{\top}\hat M^{-1}(\mathbf{H}\mathbf{e}_1)
   =\hat{\bmu}^{\top}\hat M^{-1}\hat{\bmu}.
\end{equation}
We now compute the same two quantities from \eqref{eq:kent-block}. Since
$S_T$ is positive definite, let
$s=a-\mathbf{c}^{\top}S_T^{-1}\mathbf{c}$, the Schur complement of $S_T$.
Then,
\begin{equation}\label{eq:kent-ldu}
   \begin{pmatrix} a & \mathbf{c}^{\top}\\ \mathbf{c} & S_T\end{pmatrix}
   =\begin{pmatrix} 1 & \mathbf{c}^{\top}S_T^{-1}\\ \mathbf{0} & \mathbf{I}\end{pmatrix}
    \begin{pmatrix} s & \mathbf{0}^{\top}\\ \mathbf{0} & S_T\end{pmatrix}
    \begin{pmatrix} 1 & \mathbf{0}^{\top}\\ S_T^{-1}\mathbf{c} & \mathbf{I}\end{pmatrix}.
\end{equation}
Taking determinants in \eqref{eq:kent-ldu}
and using \eqref{eq:kent-two-quantities} gives
\[
   |\hat M|=s\,|S_T|.
\]
Also,
\[
   \big(\mathbf{H}\hat M\mathbf{H}\big)^{-1} = \begin{pmatrix} a & \mathbf{c}^{\top}\\ \mathbf{c} & S_T\end{pmatrix}^{-1}
   =\begin{pmatrix} 1 & \mathbf{0}^{\top}\\ -S_T^{-1}\mathbf{c} & \mathbf{I}\end{pmatrix}
    \begin{pmatrix} s^{-1} & \mathbf{0}^{\top}\\ \mathbf{0} & S_T^{-1}\end{pmatrix}
    \begin{pmatrix} 1 & -\mathbf{c}^{\top}S_T^{-1}\\ \mathbf{0} & \mathbf{I}\end{pmatrix},
\]
so $\mathbf{e}_1^{\top}(\mathbf{H}\hat M\mathbf{H})^{-1}\mathbf{e}_1 = s^{-1}$. Therefore, by \eqref{eq:kent-two-quantities},
\[
   \hat{\bmu}^{\top}\hat M^{-1}\hat{\bmu}=s^{-1},
\]
and finally
\[
   |S_T(A)|=\frac{|\hat M(A)|}{s}
   =|\hat M(A)|\,\hat{\bmu}^{\top}\hat M(A)^{-1}\hat{\bmu}.
\]
\end{proof}

\begin{corollary}\label{cor:tn-node-score}
Under the conditions of Theorem~\ref{thm:kent-llhat}, the maximum of
$\ell(A)+J(A;\hat{\bmu})$ over $\mathbf{K}$ and $\Lambda$ is 
\begin{equation}\label{eq:tn-node-score}
   \hat\ell^{*}_{\mathrm{TN}}(A)=\hat\ell_{\mathrm{TN}}(A)+J(A),
   \qquad
   J(A)=\sum_{t\in A}\ln\bigl(\hat{\bmu}(A)^{\top}\mathbf{u}^{(t)}\bigr).
\end{equation}
\end{corollary}
 
\begin{proof}
$J(A;\hat{\bmu})$ does not depend on $\mathbf{K}$ or $\Lambda$.
\end{proof}
 
\noindent
Of the $\tfrac12k(k+1)-1$ parameters of a node,
only the $k-1$ of its mean direction are estimated for a candidate child. It is easy to see that
\begin{equation}\label{eq:kent-schur-det}
   \ln\bigl|S_T(A)\bigr|
   =\ln\Bigl\{\bigl|\hat M(A)\bigr|\,\hat{\bmu}^{\top}\hat M(A)^{-1}\hat{\bmu}\Bigr\}
   =\sum_{m=2}^{k}\ln\hat\lambda_m,
\end{equation}
where $\hat\lambda_2\ge\cdots\ge\hat\lambda_k$, the eigenvalues of $S_T(A)$,
are the fitted tangent variances. The gain does not need them: a single
Cholesky factorisation $\hat M(A)=\mathbf{L}\mathbf{L}^{\top}$ gives
$\ln|\hat M(A)|=2\sum_j\ln L_{jj}$ and
$\hat{\bmu}^{\top}\hat M(A)^{-1}\hat{\bmu}=\lVert\mathbf{L}^{-1}\hat{\bmu}\rVert^{2}$
directly from the sufficient statistics \eqref{eq:kent-suffstat}, which are
additive over the observations of a node.

\noindent
The Jacobian term, however, is not additive, and computing it exactly
requires all observations across all candidate child nodes. It can instead be
approximated from the sufficient statistics. Let
$c_t=\hat{\bmu}(A)^{\top}\mathbf{u}^{(t)}\in(0,1]$,
$R_A=\lVert\mathbf{s}_1(A)\rVert$ and $\bar c_A=R_A/n_A$, so that
$\sum_{t\in A}c_t=R_A$ and
$\sum_{t\in A}c_t^{2}=\hat{\bmu}^{\top}S_2(A)\hat{\bmu}$. Expanding
$\ln c_t$ to second order about $\bar c_A$ gives
\begin{equation}\label{eq:tn-J-expansion}
   J(A)
   =n_A\ln\bar c_A
    -\frac{\hat{\bmu}^{\top}S_2(A)\hat{\bmu}-n_A\bar c_A^{2}}{2\bar c_A^{2}}
    +\epsilon_A,
\end{equation}
where the remainder $\epsilon_A$ is of third order in $c_t-\bar c_A$, which
is of the order of the tangent variances, because
$0\le1-c_t\le\lVert\mathbf{B}_A^{\top}\mathbf{u}^{(t)}\rVert^{2}$, where
the columns of the $k\times(k-1)$ matrix $\mathbf{B}_A$ form an orthonormal
basis of $T_{\hat{\bmu}(A)}$.

\noindent
This is all that is needed to grow a tree. If the node parameters are also
of interest, let
\begin{equation}\label{eq:kent-SL}
   S_T(A)=\mathbf{W}\diag(\hat\lambda_2,\ldots,\hat\lambda_k)\,\mathbf{W}^{\top}
\end{equation}
be the spectral decomposition of $S_T(A)=\mathbf{B}_A^{\top}\hat M(A)\mathbf{B}_A$,
with $\mathbf{W}$ orthogonal. The maximisers of
Corollary~\ref{cor:tn-node-score} are then the axes
$\hat{\mathbf{K}}=\mathbf{B}_A\mathbf{W}$ and the tangent variances
$\hat\lambda_m$, and, by \eqref{eq:lambda-def}, the Kent parameters are
\begin{equation}\label{eq:kappa-closed}
   \hat\kappa(A)=\frac{1}{k-1}\sum_{m=2}^{k}\hat\lambda_m^{-1},
   \qquad
   \hat\beta_m(A)=\tfrac12\bigl(\hat\kappa(A)-\hat\lambda_m^{-1}\bigr)
\end{equation}
\citep[eq.~(22)]{ScealyWelsh2011}, which satisfy \eqref{eq:kent-constraint}
because the $\hat\lambda_m$ are positive and in descending order.

\subsubsection{What the gain responds to}\label{sec:gain-responds}
By Corollary~\ref{cor:tn-node-score} and
\eqref{eq:kent-llhat-ST}, the gain of splitting
$\Omega$ into $L$ and $R$ is
\begin{equation}\label{eq:kent-gain}
   \Delta\ell
   =
   -\frac{1}{2}\Bigl[
      n_L \ln\bigl|S_T(L)\bigr|
    + n_R \ln\bigl|S_T(R)\bigr|
    - n   \ln\bigl|S_T(\Omega)\bigr|
   \Bigr]
   +\Delta J,
   \qquad
   \Delta J=J(L)+J(R)-J(\Omega).
\end{equation}

\noindent
From \eqref{eq:kent-gain} alone, it is not immediately clear how changes in mean direction and covariance contribute to the gain. The parent and its two children are represented in their own tangent coordinates, so their covariance matrices are not directly comparable, and their mean directions are not explicitly represented in the gain. To separate these effects, we express all three nodes in a common coordinate system: the tangent coordinates of the parent.
For a node $A$, let $\bar{\mathbf{u}}_A$ and $C_A$ denote the mean and the
covariance matrix of the observations of $A$ in $\R^k$, and let
$\mathbf{B}_A$ be as in \eqref{eq:tn-J-expansion}. Since
$\bar{\mathbf{u}}_A$ is parallel to $\hat{\bmu}(A)$ and
$\mathbf{B}_A^{\top}\hat{\bmu}(A)=\mathbf{0}$, the tangent coordinates
$\mathbf{B}_A^{\top}\mathbf{u}^{(t)}$ have mean zero in $A$, and
$S_T(A)=\mathbf{B}_A^{\top}\hat M(A)\mathbf{B}_A=\mathbf{B}_A^{\top}C_A\mathbf{B}_A$
is their covariance matrix. The node mean is thus absorbed into the choice
of the tangent plane.

\noindent
Now let $\mathbf{B}=\mathbf{B}_\Omega$ and $\mathbf{w}^{(t)}=\mathbf{B}^{\top}\mathbf{u}^{(t)}$ denote the parent's tangent coordinates. The parent is centred at zero, but a child $c\in\{L,R\}$ generally has a nonzero mean
$
\bar{\mathbf{w}}_c=\mathbf{B}^{\top}\bar{\mathbf{u}}_c
$. Its covariance in the parent's coordinates is
$
\tilde S_T(c)=\mathbf{B}^{\top}C_c\mathbf{B}.
$
Thus, $S_T(c)$ and $\tilde S_T(c)$ represent the same observations projected onto the child's and parent's tangent spaces, respectively. They coincide up to a coordinate rotation when their mean directions agree, that is, when $\hat{\bmu}(c)=\hat{\bmu}(\Omega)$.

\noindent
Let $\boldsymbol{\delta}=\bar{\mathbf{w}}_L-\bar{\mathbf{w}}_R$. Then
\begin{equation}\label{eq:kent-pooled}
   nS_T(\Omega)
   =\sum_{c\in\{L,R\}}n_c\bigl\{\tilde S_T(c)+\bar{\mathbf{w}}_c\bar{\mathbf{w}}_c^{\top}\bigr\}
   =n\bar{\boldsymbol{\Sigma}}+\frac{n_Ln_R}{n}\,\boldsymbol{\delta}\boldsymbol{\delta}^{\top},
\end{equation}
where $\bar{\boldsymbol{\Sigma}}=\{n_L\tilde S_T(L)+n_R\tilde S_T(R)\}/n$. Therefore,
\begin{equation}\label{eq:kent-decomposition}
\begin{aligned}
   \Delta\ell
   ={}&\underbrace{\frac12\Bigl[n\ln\bigl|\bar{\boldsymbol{\Sigma}}\bigr|
     -\sum_{c}n_c\ln\bigl|\tilde S_T(c)\bigr|\Bigr]}_{\text{covariance}}
   +\underbrace{\frac{n}{2}\ln\Bigl(1+\frac{n_Ln_R}{n^{2}}\,
      \boldsymbol{\delta}^{\top}\bar{\boldsymbol{\Sigma}}^{-1}\boldsymbol{\delta}\Bigr)}_{\text{mean}}\\
   &+{\frac12\sum_{c}n_c\ln\frac{|\tilde S_T(c)|}{|S_T(c)|}
     +\Delta J}.
\end{aligned}
\end{equation}
The covariance term is
non-negative, because $\ln|\cdot|$ is strictly concave on positive definite
matrices, and zero only when $\tilde S_T(L)=\tilde S_T(R)$; it rewards a
change in the tangent variances or in the orientation of the axes. The mean
term is non-negative and zero only when $\boldsymbol{\delta}=\mathbf{0}$; it
rewards a shift of the mean direction, which counts more when
$\boldsymbol{\delta}$ points along a direction of small tangent variance.
The gain therefore responds to both the covariance and the mean.

\noindent
The last two terms arise from expressing the child nodes in their own tangent coordinates rather than in the parent's coordinates. We express all three nodes in the parent's tangent coordinates to make the contributions to the split gain easier to interpret. This change of coordinates is closely related to the role of the Jacobian term, which ensures that the node log-likelihoods are densities with respect to a common measure \eqref{eq:kent-working-loglik}. We do not analyse these terms further.

\subsection{Logistic-normal}\label{sec:mln}\label{sec:mln-computation}

A composition $\mathbf{Z}$ follows the multivariate logistic-normal
distribution, $\mathbf{Z}\sim\mathrm{MLN}(\boldsymbol{\mu},\boldsymbol{\Sigma})$,
if the additive log-ratio vector $\mathbf{y}=(y_1,\ldots,y_{k-1})^{\top}$,
$y_j=\ln(z_j/z_k)$, is $N_{k-1}(\boldsymbol{\mu},\boldsymbol{\Sigma})$
\citep{RefArt_MLNdist}. On $\mathbf{S}^k$, its density is
\begin{equation}\label{eq:mln-density}
   f(\mathbf{z};\boldsymbol{\mu},\boldsymbol{\Sigma})
   =\varphi_{\boldsymbol{\Sigma}}(\mathbf{y}-\boldsymbol{\mu})
    \prod_{j=1}^{k}z_j^{-1}.
\end{equation}
For the vectors $\mathbf{y}^{(t)}$, $t\in A$, of a node $A$, the maximum
likelihood estimators are 
\begin{equation}\label{eq:suffstat}
   \hat{\boldsymbol{\mu}}(A)=\frac{\mathbf{s}_1(A)}{n_A},
   \qquad
   \hat{\boldsymbol{\Sigma}}(A)
   =\frac{1}{n_A}\sum_{t\in A}(\mathbf{y}^{(t)}-\hat{\boldsymbol{\mu}}(A))(\mathbf{y}^{(t)}-\hat{\boldsymbol{\mu}}(A))^{\top}
   =\frac{S_2(A)}{n_A}-\hat{\boldsymbol{\mu}}(A)\hat{\boldsymbol{\mu}}(A)^{\top},
\end{equation}
where $\mathbf{s}_1(A)=\sum_{t\in A}\mathbf{y}^{(t)}$ and
$S_2(A)=\sum_{t\in A}\mathbf{y}^{(t)}\mathbf{y}^{(t)\top}$ are additive
sufficient statistics. The gain \eqref{eq:general-split} of a split
$\Omega\to(L,R)$ is
\begin{equation}\label{eq:mln-gain}
   \Delta\ell
   =-\frac12\Bigl[
      n_L\ln\bigl|\hat{\boldsymbol{\Sigma}}(L)\bigr|
     +n_R\ln\bigl|\hat{\boldsymbol{\Sigma}}(R)\bigr|
     -n\ln\bigl|\hat{\boldsymbol{\Sigma}}(\Omega)\bigr|
   \Bigr].
\end{equation}
This gain has no separate location term either. As in
\eqref{eq:kent-pooled}, 
$\hat{\boldsymbol{\Sigma}}(\Omega)=\bar{\boldsymbol{\Sigma}}
+(n_Ln_R/n^{2})\,\boldsymbol{\delta}\boldsymbol{\delta}^{\top}$, where 
$\boldsymbol{\delta}=\hat{\boldsymbol{\mu}}(L)-\hat{\boldsymbol{\mu}}(R)$ and
$\bar{\boldsymbol{\Sigma}}=\{n_L\hat{\boldsymbol{\Sigma}}(L)+n_R\hat{\boldsymbol{\Sigma}}(R)\}/n$,
and, as in \eqref{eq:kent-decomposition},
\begin{equation}\label{eq:mln-decomposition}
\begin{aligned}
   \Delta\ell
   ={}&\underbrace{\frac12\Bigl[
      n\ln\bigl|\bar{\boldsymbol{\Sigma}}\bigr|
     -\sum_{c}n_c\ln\bigl|\hat{\boldsymbol{\Sigma}}(c)\bigr|
   \Bigr]}_{\text{covariance}}\\
   &+\underbrace{\frac{n}{2}\ln\Bigl(1+\frac{n_Ln_R}{n^{2}}\,
      \boldsymbol{\delta}^{\top}\bar{\boldsymbol{\Sigma}}^{-1}\boldsymbol{\delta}\Bigr)}_{\text{mean}}.
\end{aligned}
\end{equation}
This is the
classical factorisation of the likelihood ratio for the equality of two
normal distributions \citep[Ch.~10]{Anderson2003}: the first term tests
equal covariance matrices and rewards a change in the variances or in the
dependence, and the second, an increasing function of Hotelling's
two-sample statistic, tests equal means and rewards a shift of the mean.

\noindent
Like the TN distribution, the MLN distribution has $k-1$ parameters in its
mean and $\tfrac12k(k-1)$ in its covariance, $\tfrac12k(k+1)-1$ in all. Both
models are based on the normal distribution, and the Gaussian parts of
their gains, \eqref{eq:mln-gain} and \eqref{eq:kent-gain}, have the same
form and the same decomposition, \eqref{eq:mln-decomposition} and
\eqref{eq:kent-decomposition}. The difference lies in the coordinates.
Unlike the tangent coordinates, the log-ratio coordinates are the same in
every node, so the Jacobian $\prod_jz_j^{-1}$ in \eqref{eq:mln-density}
does not depend on the parameters and cancels from the gain. The tangent
coordinates, in contrast, are defined by the node's own mean direction, so
their Jacobian does not cancel, and the TN gain has the additional location
term $\Delta J$. The two models also behave differently near the boundary of
the simplex: the logarithm of a part close to zero is highly variable, and
its variance then dominates $\hat{\boldsymbol{\Sigma}}$, whereas the square
root stays bounded and is defined at zero.

\section{Simulation study}\label{Sec_results}

\noindent
The Dirichlet random forest has been compared with other random forests for
compositional responses in two recent studies
\citep{Khaled_Steohan_2026,Stephan_Khaled_2026}. Rather than repeating these
comparisons, we focus on comparing the three split criteria introduced in
Section~\ref{sec:nodes}, Dirichlet, logistic-normal and TN, across a range
of simulated designs.

\subsection{Designs and evaluation}\label{sec:sim-design}

\noindent
Consider a sample of size $n$, $(\mathbf{x}_i,\mathbf{z}_i)$,
$i=1,\ldots,n$, where $\mathbf{x}_i=(x_{i1},\ldots,x_{i10})^{\top}$ has
independent components uniformly distributed on $(-1,1)$, and
$\mathbf{z}_i=(z_{i1},z_{i2},z_{i3})^{\top}$ is a composition drawn from a
conditional distribution given $\mathbf{x}_i$. In every design, $x_{i1}$ and
$x_{i2}$ affect only the location of this distribution, $x_{i3}$ and
$x_{i4}$ only its shape (spread or dependence), $x_{i5}$ both, and
$x_{i6},\ldots,x_{i10}$ are noise variables; \ref{app:sim} gives
the full specifications. In design D, $\mathbf{z}_i$ follows a Dirichlet
distribution. In design L, the additive log-ratios of $\mathbf{z}_i$ follow
a bivariate normal distribution with a correlation of approximately $-0.9$,
and the location varies mainly along the direction of smallest variance. In
design K, $\sqrt{\mathbf{z}_i}$ follows an exact Kent distribution, one part
is close to zero, and the location varies mainly along the direction of
smallest variance.

\noindent All three forests, called Dir-RF, MLN-RF and TN-RF, use $1000$
trees and $m_{\mathrm{try}}=3$ candidate covariates per split, grow every
tree on all training observations, and predict with the forest-weighted mean
$\hat{\mathbf{z}}(\mathbf{x})=\sum_{t}w_t(\mathbf{x})\,\mathbf{z}^{(t)}$,
with the weights \eqref{eq:forest-weights}; TN-RF computes the Jacobian term
by \eqref{eq:tn-J-expansion}. The three forests use the same growth rule. A
node is split only while its depth is less than $10$, and only into two
children that each contain at least $c=\max(m,k+2)$ observations, where $m$
is the minimum child size, so a node with fewer than $2c$ observations is
not split. This rule differs from the default behaviour of common random
forest libraries, such as \texttt{randomForest} and \texttt{ranger}, in
which the node-size parameter decides whether a node may be split, but a
split may still produce a much smaller child. Our split criteria compare
the maximum likelihoods of the children, and the likelihood of a small
child is unstable, because its $(k-1)\times(k-1)$ covariance matrix is
estimated from few observations; the floor $k+2$ keeps this estimate
non-singular.

\noindent All reported numbers come from independent runs with $R=40$
replications: in each, $n$ observations are drawn, $80\%$ are used for
training and $20\%$ for testing, and the three forests see the same data, so
all comparisons are paired. The designs are run at
$n\in\{200,500,1000,2000\}$. As in \citet{Khaled_Steohan_2026}, accuracy on
the test set $\mathcal{T}$ is measured by the Aitchison distance
\citep{RefArt_logratiocompdistance} and the averaged root mean squared
error,
\begin{gather}
   \mathrm{AD}=\frac{1}{|\mathcal{T}|}\sum_{i\in\mathcal{T}}
   \bigl\lVert\operatorname{clr}\mathbf{z}_i
   -\operatorname{clr}\hat{\mathbf{z}}(\mathbf{x}_i)\bigr\rVert,
   \label{eq:aitchison-measure}\\
   \text{RMSE-averaged}=\frac1k\sum_{j=1}^{k}
   \Bigl[\frac{1}{|\mathcal{T}|}\sum_{i\in\mathcal{T}}
   \bigl\{z_{ij}-\hat z_j(\mathbf{x}_i)\bigr\}^{2}\Bigr]^{1/2},
   \label{eq:rmse-measure}
\end{gather}
where $\operatorname{clr}\mathbf{z}=(\ln z_1-\bar g,\ldots,\ln z_k-\bar g)$,
with $\bar g=k^{-1}\sum_{l}\ln z_l$, is the centred log-ratio transform.

\subsection{Prediction accuracy}\label{sec:res-accuracy}

\noindent
Table~\ref{tab:sample-size} reports the test errors of the full designs at
four sample sizes. In design D, TN-RF had the lowest Aitchison distance and
RMSE-averaged at every sample size, and the differences were significant
when $n\ge 1000$. In design L, MLN-RF had the lowest RMSE-averaged at every
sample size and the lowest Aitchison distance when $n\ge 500$, and the
differences on both measures were significant when $n\ge 1000$. In design K,
TN-RF had the lowest Aitchison distance at every sample size, significantly
from $n=500$ on; it also had the lowest RMSE-averaged when $n\ge 500$, but
these differences were not significant. Dir-RF was the least accurate forest
in L at every sample size, and it had the lowest error only once, for
RMSE-averaged in K at $n=200$.

\begin{table}[h!]
\centering
\caption{Test error at four sample sizes with minimum child size $m=10$.
Entries are means over $40$ paired replications, with standard deviations
in parentheses; smaller is better. Bold: lowest mean for that design and
$n$.}
\label{tab:sample-size}
\small
\begin{tabular}{rl ccc}
\toprule
$n$ & Forest & Design D & Design L & Design K \\
\midrule
\multicolumn{5}{l}{\emph{Aitchison distance}} \\
200 & Dir-RF & 0.5078 (0.0608) & 1.1576 (0.1277) & 0.7919 (0.1674) \\
 & MLN-RF & 0.5062 (0.0625) & 1.1488 (0.1356) & 0.7969 (0.1656) \\
 & TN-RF & \textbf{0.5049} (0.0625) & \textbf{1.1480} (0.1354) & \textbf{0.7894} (0.1653) \\[3pt]
500 & Dir-RF & 0.4710 (0.0356) & 1.2331 (0.1001) & 0.6902 (0.0911) \\
 & MLN-RF & 0.4711 (0.0348) & \textbf{1.2237} (0.0993) & 0.6899 (0.0911) \\
 & TN-RF & \textbf{0.4699} (0.0341) & 1.2256 (0.0987) & $\mathbf{0.6832}$ (0.0901) \\[3pt]
1000 & Dir-RF & 0.4497 (0.0215) & 1.1856 (0.0594) & 0.7200 (0.0540) \\
 & MLN-RF & 0.4483 (0.0209) & $\mathbf{1.1785}$ (0.0597) & 0.7196 (0.0535) \\
 & TN-RF & $\mathbf{0.4471}$ (0.0207) & 1.1806 (0.0587) & $\mathbf{0.7128}$ (0.0525) \\[3pt]
2000 & Dir-RF & 0.4368 (0.0128) & 1.1970 (0.0365) & 0.7072 (0.0364) \\
 & MLN-RF & 0.4369 (0.0123) & $\mathbf{1.1905}$ (0.0376) & 0.7051 (0.0361) \\
 & TN-RF & $\mathbf{0.4358}$ (0.0124) & 1.1922 (0.0368) & $\mathbf{0.6994}$ (0.0362) \\
\midrule
\multicolumn{5}{l}{\emph{RMSE-averaged} ($\times100$)} \\
200 & Dir-RF & 9.811 (0.79) & 19.453 (1.49) & \textbf{5.584} (0.60) \\
 & MLN-RF & 9.774 (0.80) & \textbf{19.162} (1.60) & 5.607 (0.61) \\
 & TN-RF & \textbf{9.736} (0.80) & 19.181 (1.62) & 5.598 (0.59) \\[3pt]
500 & Dir-RF & 9.213 (0.68) & 20.098 (1.14) & 5.389 (0.33) \\
 & MLN-RF & 9.215 (0.67) & $\mathbf{19.762}$ (1.11) & 5.381 (0.34) \\
 & TN-RF & \textbf{9.190} (0.66) & 19.847 (1.11) & \textbf{5.379} (0.34) \\[3pt]
1000 & Dir-RF & 8.822 (0.35) & 19.590 (0.65) & 5.460 (0.27) \\
 & MLN-RF & 8.799 (0.34) & $\mathbf{19.344}$ (0.65) & 5.443 (0.26) \\
 & TN-RF & $\mathbf{8.771}$ (0.33) & 19.412 (0.63) & \textbf{5.441} (0.26) \\[3pt]
2000 & Dir-RF & 8.561 (0.27) & 19.638 (0.38) & 5.444 (0.20) \\
 & MLN-RF & 8.559 (0.26) & $\mathbf{19.421}$ (0.40) & 5.435 (0.20) \\
 & TN-RF & $\mathbf{8.539}$ (0.26) & 19.472 (0.40) & \textbf{5.430} (0.20) \\
\bottomrule
\end{tabular}
\end{table}

\noindent
TN-RF was the most accurate forest in two designs whose data do not follow
its node model, and it was never the least accurate of the three forests in
Table~\ref{tab:sample-size}. In K, the square root of the composition
follows a Kent distribution, which the TN distribution only approximates
(Section~\ref{sec:tn-kent}), and one part is close to zero: $30\%$ of the
draws have a part below $0.01$. In D, the compositions are Dirichlet, so
the node model of Dir-RF is the correct one. Nevertheless, TN-RF had a lower
error than Dir-RF at every sample size and on both measures. These
differences were small, below $1\%$ of the error, but at $n=1000$ and
$n=2000$ they were significant against the second-best forest.

\section{Conclusion}\label{Sec_concl}

\noindent
We include all three criteria so that the forest can be used within any of the three approaches.

\noindent The three criteria correspond to the three spaces in which
compositions are usually analysed. On the simplex itself, we used the
Dirichlet criterion, which was developed in earlier work
\citep{Khaled_Steohan_2026,Stephan_Khaled_2026}. In log-ratio coordinates,
we used the MLN criterion, whose maximum log-likelihood has a closed-form
expression based on additive sufficient statistics. For square-root
coordinates on the sphere, we defined the TN distribution for computational
reasons.

\noindent The criteria differ in what they respond to. The Dirichlet
criterion judges a shift of the mean approximately as if the variance of
the square-root coordinates were the same in every direction
\eqref{eq:dir-isotropic}. The MLN and TN criteria judge it relative to the
full covariance, and they also respond to a change in the covariance, in
log-ratio and in tangent coordinates, respectively. The two normal models
trade different properties: the MLN distribution is defined on a space that
is in one-to-one correspondence with the interior of the simplex, but
requires strictly positive parts, whereas the square-root coordinates of the
TN distribution stay bounded and are defined at zero, at the price of some
probability outside the positive orthant. 

\noindent In our simulations, TN-RF was the most accurate of the three
forests in two of the three designs and, in the full designs, never the
least accurate. 

\bibliography{References.bib}
\printnoidxglossary[type=acronym] 

\appendix

\section{Simulation designs and additional results}
\label{app:sim}

In every design, the location enters through the score
\begin{equation}\label{eq:sim-location}
   \eta(\mathbf{x})=\sin(\pi x_1)+1.5\,(x_2^{2}-\tfrac13)+0.5\,x_5,
\end{equation}
and the shape through $s_3(\mathbf{x})=\tanh(2x_3)$,
$s_4(\mathbf{x})=\tfrac12\operatorname{sign}(x_4)$ and $x_5$. We use
$\mathcal{C}$ for the closure operation, which divides a positive vector by
the sum of its entries.

\paragraph{Design D (Dirichlet)} The response is
\begin{equation}\label{eq:sim-D}
   \mathbf{Z}\mid\mathbf{x}\sim\operatorname{Dir}\bigl(\alpha_0(\mathbf{x})\,\mathbf{m}(\mathbf{x})\bigr),
   \qquad
   \mathbf{m}(\mathbf{x})=\mathcal{C}\bigl(e^{0.8\sin(\pi x_1)+0.5x_5},\,e^{1.2(x_2^{2}-1/3)},\,1\bigr),
\end{equation}
with $\alpha_0(\mathbf{x})=40\exp\{0.6\,s_3(\mathbf{x})+0.6\,s_4(\mathbf{x})+0.4\,x_5\}$.
The shape covariates change only the precision, so
$E(\mathbf{Z}\mid\mathbf{x})=\mathbf{m}(\mathbf{x})$ does not depend on them,
and, by \eqref{eq:dir-isotropic}, the covariance of $\sqrt{\mathbf{Z}}$ is
approximately isotropic.

\paragraph{Design L (logistic-normal)} The additive log-ratios
$\mathbf{y}=(\ln Z_1/Z_3,\ln Z_2/Z_3)^{\top}$ are
\begin{equation}\label{eq:sim-L}
   \mathbf{y}\mid\mathbf{x}\sim N_2\bigl(\bmu(\mathbf{x}),\Sigma(\mathbf{x})\bigr),
   \qquad
   \Sigma(\mathbf{x})=\sigma^{2}(\mathbf{x})
   \begin{pmatrix}1&\rho(\mathbf{x})\\ \rho(\mathbf{x})&1\end{pmatrix},
\end{equation}
with $\rho(\mathbf{x})=-0.9+0.05\,s_3(\mathbf{x})$ and
$\sigma^{2}(\mathbf{x})=\exp\{0.5\,s_4(\mathbf{x})+0.3\,x_5\}$. Because
$\rho<0$, the minor axis of $\Sigma$ is $\mathbf{e}_{\min}=(1,1)^{\top}/\sqrt2$,
with variance $\sigma^{2}(1+\rho)\approx0.1\,\sigma^{2}$, and the major axis
is $\mathbf{e}_{\max}=(1,-1)^{\top}/\sqrt2$, with variance
$\sigma^{2}(1-\rho)\approx1.9\,\sigma^{2}$. The mean moves mainly along the
minor axis,
\[
   \bmu(\mathbf{x})=1.5\,(0.1)^{1/2}\,\eta(\mathbf{x})\,
   \bigl\{\cos(\pi/8)\,\mathbf{e}_{\min}+\sin(\pi/8)\,\mathbf{e}_{\max}\bigr\},
\]
so that one unit of $\eta$ moves the mean by about $1.5$ minor-axis standard
deviations. A Dirichlet distribution cannot represent this dependence.

\paragraph{Design K (Kent)} The square root of the composition follows an
exact Kent distribution. Let 
$\bmu_0=\sqrt{\mathcal{C}(1,1,0.040)}$. At a direction $\bmu$, set
$\mathbf{s}(\bmu)=(\mathbf{I}-\bmu\bmu^{\top})\mathbf{e}_3/\|(\mathbf{I}-\bmu\bmu^{\top})\mathbf{e}_3\|$
and
$\mathbf{l}(\bmu)=\mathbf{s}(\bmu)\times\bmu$. Given $\mathbf{x}$,
$\mathbf{U}$ has the Kent density \eqref{eq:kent-density} with $k=3$,
\begin{equation}\label{eq:sim-K}
   f(\mathbf{u}\mid\mathbf{x})\propto
   \exp\Bigl[\kappa(\mathbf{x})\,\bmu(\mathbf{x})^{\top}\mathbf{u}
   +\beta(\mathbf{x})\bigl\{(\mathbf{l}^{\top}\mathbf{u})^{2}-(\mathbf{s}^{\top}\mathbf{u})^{2}\bigr\}\Bigr],
\end{equation}
where $\mathbf{l}=\mathbf{l}(\bmu(\mathbf{x}))$,
$\mathbf{s}=\mathbf{s}(\bmu(\mathbf{x}))$,
$\kappa(\mathbf{x})=330\exp\{0.5\,s_4(\mathbf{x})+0.3\,x_5\}$ and
$\beta(\mathbf{x})=d(\mathbf{x})\kappa(\mathbf{x})$. Also,
\[
   \bmu(\mathbf{x})\propto\bmu_0+1.5\,\bar\lambda_{\mathrm{s}}^{1/2}\,\eta(\mathbf{x})\,
   \bigl\{\cos(3\pi/16)\,\mathbf{s}(\bmu_0)+\sin(3\pi/16)\,\mathbf{l}(\bmu_0)\bigr\},
\]
normalised to unit length, where $\bar\lambda_{\mathrm{s}}=1/(330\times1.4)$
is the short-axis variance at the base values.

\begin{table}[h!]
\centering
\caption{Test error at $n=1000$ with minimum child size $m=10$ and $m=20$.
Entries are means over $40$ paired replications. Bold: lowest mean in the
row for that measure.}
\label{tab:m-sensitivity}
\small
\begin{tabular}{ll ccc ccc}
\toprule
 & & \multicolumn{3}{c}{Aitchison distance} & \multicolumn{3}{c}{RMSE-averaged ($\times100$)} \\
\cmidrule(lr){3-5}\cmidrule(lr){6-8}
Design & $m$ & Dir & MLN & TN & Dir & MLN & TN \\
\midrule
D & 10 & 0.4497 & 0.4483 & $\mathbf{0.4471}$ & 8.822 & 8.799 & $\mathbf{8.771}$ \\
 & 20 & 0.4644 & 0.4631 & $\mathbf{0.4617}$ & 9.099 & 9.075 & $\mathbf{9.044}$ \\
\midrule
L & 10 & 1.1856 & $\mathbf{1.1785}$ & 1.1806 & 19.590 & $\mathbf{19.344}$ & 19.412 \\
 & 20 & 1.1842 & $\mathbf{1.1779}$ & 1.1795 & 19.520 & $\mathbf{19.311}$ & 19.357 \\
\midrule
K & 10 & 0.7200 & 0.7196 & $\mathbf{0.7128}$ & 5.460 & 5.443 & \textbf{5.441} \\
 & 20 & 0.7308 & 0.7315 & $\mathbf{0.7243}$ & 5.460 & 5.458 & \textbf{5.451} \\
\bottomrule
\end{tabular}
\end{table}

\newpage
\section{Normalising constants of the TN and Kent distributions}\label{app:constants}

We compared the constants that are omitted when the log-likelihood of each
distribution is replaced by the Gaussian log-likelihood of its tangent
coordinates. The settings were $k=3,4,5$, matched parameters
\eqref{eq:lambda-def}, $d_m=\beta_m/\kappa$ equally spaced from $d$ to $-d$
with $d\in\{0,0.1,0.2,0.3,0.4,0.45\}$, and $2\le\kappa\le1000$.

 For the TN distribution, the omitted constant
is $-\ln\pi_\Lambda$, computed from the series of \citet{Ruben1962}. For the
Kent distribution, it is $\ln c_G-\ln c(\kappa,\bbeta)$, where
$c_G=e^{\kappa}(2\pi)^{(k-1)/2}|\Lambda|^{1/2}$ and $c(\kappa,\bbeta)$ was
computed by nested adaptive quadrature. Both agreed with closed forms and
with independent numerical methods to within $10^{-12}$.

 The TN constant decays as
$\exp\{-\kappa(1-2d_2)/2\}$, up to a factor polynomial in $\kappa$, and
exceeded $10^{-3}$ nats only when $\kappa(1-2d_2)\le15$. The Kent constant
decays as $\kappa^{-1}\{(a_1^{2}+2a_2)/8-a_1/2\}$, with
$a_r=\sum_m(1-2d_m)^{-r}$. For $k=3$ and
$d_2=-d_3=0.2$, it exceeded $10^{-3}$ nats up to $\kappa=330$, and for $k=5$
over the whole range studied.

\end{document}